# AllMetrics: A Unified Python Library for Standardized Metric Evaluation and Robust Data Validation in Machine Learning

Morteza Alizadeh[1], Mehrdad Oveisi[2,3], Sonya Falahati[3,4], Ghazal Mousavi[5], Mohsen Alambardar Meybodi[6], Somayeh Sadat Mehrnia[7], Ilker Hacihaliloglu[8,9], Arman Rahmim[9,10], Mohammad R. Salmanpour[3,9,10*]

[1] Department of Mathematics, University of Isfahan, Isfahan, Iran
[2] Department of Computer Science, University of British Columbia, Vancouver, BC, Canada
[3] Technological Virtual Collaboration (TECVICO Corp.), Vancouver, BC, Canada
[4] Electrical and Computer Engineering Department, Nooshirvani University of Technology, Babol, Iran
[5] School of Electrical and Computer Engineering, University of Tehran, Tehran, Iran
[6] Department of Applied Mathematics and Computer Science, University of Isfahan, Isfahan, Iran
[7] Department of Integrative Oncology, Breast Cancer Research Center, Motamed Cancer Institute, ACECR, Tehran, Iran
[8] Department of Medicine, University of British Columbia, Vancouver, BC, Canada
[9] Department of Radiology, University of British Columbia, Vancouver, BC, Canada
[10] Department of Integrative Oncology, BC Cancer Research Institute, Vancouver, BC, Canada

(*) Corresponding Author: Mohammad R. Salmanpour, PhD (*msalman@bccrc.ca*)

**ABSTRACT**
Machine learning (ML) models rely heavily on consistent and accurate performance metrics to evaluate and compare their effectiveness. However, existing libraries often suffer from fragmentation, inconsistent implementations, and insufficient data validation protocols, leading to unreliable results. Existing libraries have often been developed independently and without adherence to a unified standard, particularly concerning the specific tasks they aim to support. As a result, each library tends to adopt its conventions for metric computation, input/output formatting, error handling, and data validation protocols. This lack of standardization leads to both implementation differences (ID) and reporting differences (RD), making it difficult to compare results across frameworks or ensure reliable evaluations. To address these issues, we introduce AllMetrics, an open-source unified Python library designed to standardize metric evaluation across diverse ML tasks, including regression, classification, clustering, segmentation, and image-to-image translation. The library implements class-specific reporting for multi-class tasks through configurable parameters to cover all use cases, while incorporating task-specific parameters to resolve metric computation discrepancies across implementations. Various datasets from domains like healthcare, finance, and real estate were applied to our library and compared with Python, Matlab, and R components to identify which yield similar results. AllMetrics combines a modular Application Programming Interface (API) with robust input validation mechanisms to ensure reproducibility and reliability in model evaluation. This paper presents the design principles, architectural components, and empirical analyses demonstrating the ability to mitigate evaluation errors and to enhance the trustworthiness of ML workflows.

## 1. INTRODUCTION

Machine learning (ML) models fundamentally rely on quantitative metrics to assess performance across essential tasks, including classification, regression, and clustering. These metrics provide crucial insights into model behavior that enable meaningful comparisons, systematic optimizations, and successful deployments. However, the current utility of these metrics is significantly undermined by widespread inconsistencies in their computation and reporting, frequently leading to unreliable results. Additional challenges, including missing values, imbalanced class distributions, and mismatched dimensionalities, further exacerbate these issues, particularly when existing libraries fail to implement rigorous input validation protocols.

Recent research has identified [1] two primary sources of inconsistency plaguing ML evaluation methodologies. First, reporting differences (RD) emerge when metric documentation varies substantially across platforms, creating potential for misinterpretation - a problem particularly evident when comparing implementations between Python and R ecosystems. Second, implementation differences (ID) occur when the same nominal metric employs varying algorithmic assumptions or computational methodologies, producing divergent results even when applied to identical datasets. These inconsistencies collectively pose substantial barriers to reproducible research, reliable benchmarking, and cross-platform deployment of ML solutions.

The current landscape of evaluation tools reveals several limitations. Domain-specific libraries, while valuable within their niches, demonstrate limited interoperability. For instance, Scikit-Learn [2] serves traditional ML needs, TorchMetrics [3] specializes in deep learning applications, and Scikit-Image [4] focuses on computer vision tasks - each with its conventions and limitations. General-purpose numerical tools like SciPy [5] and computer vision libraries like OpenCV [6] provide foundational utilities but lack dedicated support for comprehensive metric validation. Language-specific solutions such as R's caret package [7] improve consistency within their respective ecosystems but remain fundamentally isolated from broader toolchains. While specialized solutions like PyCM [8] for classification or TensorFlow's [9] metrics module advance their respective domains, no existing framework successfully unifies metric evaluation with robust data validation across the complete ML lifecycle.

To address these critical gaps, this study introduces AllMetrics, a Python library specifically designed to bridge current limitations through three key innovations: (i) The library establishes unified metric computation through standardized implementations spanning regression, classification, clustering, segmentation, and image-to-image tasks; (ii) it incorporates robust data validation via automated checks for edge cases, including empty masks, class imbalances, and spatial mismatches, thereby ensuring input integrity; and (iii) the system's modular design promotes both reproducibility and practical integration through extensible, task-agnostic components.

AllMetrics systematically addresses both RD and ID issues, effectively mitigating errors that arise from fragmented tool ecosystems while significantly enhancing trust in model evaluations. Its validation framework represents a substantial advance over prior work, enforcing rigorous input criteria before metric computation - for example, properly handling missing values in regression tasks or non-binary masks in segmentation scenarios. The library makes three principal contributions to the field: a standardized API that eliminates implementation ambiguities across ML tasks; integrated data validation that proactively detects and resolves input anomalies; and empirical analysis revealing inconsistencies in existing libraries that motivate AllMetrics' design.

## 2. PROGRAMMING PREPARATION CRITERIA

AllMetrics employs a modular architectural design to provide task-specific metric evaluation while maintaining lightweight adaptability and extensibility. The library's structure enables straightforward installation through the Python Package Index (PyPI) using standard package management commands, followed by intuitive access to its comprehensive metric evaluation capabilities, using the following command: "*pip install allmetrics*"

### 2.1 Metric organization and accessibility

The library categorizes metrics according to fundamental ML tasks, each implemented as an independent module. The regression module includes common evaluation measures such as mean absolute error, mean squared error, R Squared ($R^2$; adjusted, by defining a proper parameter), and specialized loss functions including Huber Loss and Log Cosh Loss. Classification metrics encompass accuracy scores, precision-recall measures, F Scores, and probabilistic assessments like log loss. Clustering evaluation incorporates similarity indices (Rand Score, Adjusted Mutual Information) and quality measures (Silhouette Score, Davies-Bouldin Index). For segmentation tasks, the library provides spatial similarity metrics, including Dice Score and Intersection-over-Union (IoU), while image-to-image translation assessment utilizes Structural Similarity Index (SSIM) and Peak Signal-to-Noise Ratios (PSNR).

Users can programmatically discover available metrics through the "*list_of_metrics()*" method within each module. Detailed metric documentation, including parameter specifications and usage guidelines, is accessible via the "*get_metric_details()*" function. This discovery mechanism ensures transparency and facilitates proper metric selection for specific evaluation scenarios. Table 1 provides a comprehensive comparison of metrics implemented in AllMetrics alongside their availability in other popular libraries such as Scikit-Learn, PyTorch, TensorFlow, and Scikit-Image.

### 2.2 Data validation framework

The library's robust validation system represents a core architectural innovation, implemented through specialized validator classes for each task type. These validators combine shared foundational checks with task-specific validation protocols to ensure comprehensive data integrity assessment.

#### 2.2.1 Core validation functions

All validators implement a common set of fundamental checks that address universal data quality concerns. The framework verifies data types and structures, checking for appropriate array formats and consistent dimensionalities across inputs. It systematically detects missing or infinite values that could compromise computations, while length consistency checks prevent dimension mismatch errors. Statistical validations include variance analysis to prevent degenerate cases in metrics like $R^2$, non-negativity enforcement for logarithmic measures, and outlier detection through statistical scoring methods. For large datasets, the framework employs optimized sampling strategies to maintain validation efficiency without sacrificing thoroughness.

#### 2.2.2 Task-specific validation extensions

Beyond these shared capabilities, each validator implements domain-specific checks tailored to its evaluation context. The regression validator performs multicollinearity analysis to identify problematic feature correlations. Classification validators include class label consistency verification and probability distribution validation, along with class imbalance detection. Clustering evaluation incorporates feature dimensionality assessment and cluster separation analysis. For segmentation tasks, the validator checks for empty masks, validates binary input requirements, and performs spatial consistency verification. The image translation validator handles pixel value normalization and channel consistency checks across input images.

**2.2.3 Comprehensive validation interface**

The framework provides a unified "*validate_all()*" method that executes all relevant checks through a single interface, while maintaining flexibility through configurable check activation. This design allows users to balance thoroughness with computational efficiency, particularly important when working with large datasets or real-time evaluation scenarios. While the built-in core and task-specific validators cover a wide range of common data quality issues and task-specific requirements, the validation system's modular architecture supports straightforward extension, enabling researchers to incorporate custom validation rules for specialized applications.

**Table 1**. Comparison of evaluation metrics available in AllMetrics versus other major ML libraries. Checkmarks (✓) indicate full support, while dashes (–) denote no native implementation. SSIM: Structural Similarity Index Measure, PSNR: Peak Signal-to-Noise Ratio.

| Task | Metric | AllMetrics | Scikit-Learn | Pytorch | Tensorflow | Scikit-Image | PyCM |
|---|---|---|---|---|---|---|---|
| Regression | Mean Absolute Error | ✓ | ✓ | ✓ | ✓ | ✓ | - |
| | Mean Squared Error | ✓ | ✓ | ✓ | ✓ | ✓ | ✓ |
| | Mean Absolute Percentage Error | ✓ | ✓ | - | - | - | - |
| | Mean Squared Log Error | ✓ | ✓ | ✓ | ✓ | - | ✓ |
| | Mean Bias Deviation | ✓ | - | - | - | - | - |
| | Median Absolute Error | ✓ | ✓ | - | - | - | - |
| | Symmetric Mean Absolute Percentage Error | ✓ | - | - | - | - | - |
| | Relative Squared Error | ✓ | - | - | - | - | - |
| | R Squared (R²) | ✓ | ✓ | ✓ | - | - | - |
| | R Squared (R²; Adjusted) | ✓ | - | - | - | - | - |
| | Explained Variance | ✓ | ✓ | ✓ | - | - | - |
| | Huber Loss | ✓ | - | ✓ | ✓ | - | - |
| | Log Cosh Loss | ✓ | - | ✓ | ✓ | - | - |
| | Max Error | ✓ | ✓ | ✓ | ✓ | - | - |
| | Mean Tweedie Deviance | ✓ | ✓ | - | - | - | - |
| | Mean Pinball Loss | ✓ | ✓ | - | - | - | - |
| Classification | Accuracy Score | ✓ | ✓ | ✓ | ✓ | - | ✓ |
| | Precision Score | ✓ | ✓ | ✓ | ✓ | - | ✓ |
| | Recall Score | ✓ | ✓ | ✓ | ✓ | - | ✓ |
| | F1 Score | ✓ | ✓ | ✓ | ✓ | - | ✓ |
| | Balanced Accuracy | ✓ | ✓ | - | - | - | - |
| | Matthews Correlation Coefficient | ✓ | ✓ | - | - | - | ✓ |
| | Cohens Kappa | ✓ | ✓ | - | - | - | ✓ |
| | F Beta Score | ✓ | ✓ | - | - | - | - |
| | Jaccard Score | ✓ | ✓ | - | ✓ | - | ✓ |
| | Hamming Loss | ✓ | ✓ | - | - | - | - |
| | Log Loss | ✓ | ✓ | - | ✓ | - | - |
| | Confusion Matrix | ✓ | ✓ | ✓ | ✓ | - | ✓ |
| | Top K Accuracy | ✓ | ✓ | ✓ | ✓ | - | - |
| Clustering | Rand Score | ✓ | ✓ | - | - | - | - |
| | Adjusted Rand Index | ✓ | ✓ | - | - | - | - |
| | Mutual Info Score | ✓ | ✓ | - | - | - | - |
| | Normalized Mutual Info Score | ✓ | ✓ | - | - | - | - |
| | Adjusted Mutual Info Score | ✓ | ✓ | - | - | - | - |
| | Silhouette Score | ✓ | ✓ | - | - | - | - |
| | Calinski Harabasz Index | ✓ | ✓ | - | - | - | - |
| | Davies-Bouldin Index | ✓ | ✓ | - | - | - | - |
| | Homogeneity Score | ✓ | ✓ | - | - | - | - |
| | Completeness Score | ✓ | ✓ | - | - | - | - |
| | V Measure Score | ✓ | ✓ | - | - | - | - |
| | Fowlkes-Mallows Score | ✓ | ✓ | - | - | - | - |
| Segmentation | Dice Score | ✓ | ✓ | ✓ | ✓ | - | - |
| | IoU Score | ✓ | - | - | - | - | - |
| | Sensitivity | ✓ | ✓ | - | - | - | - |
| | Specificity | ✓ | ✓ | - | - | - | - |
| | Precision | ✓ | ✓ | - | - | - | - |
| | Hausdorff Distance | ✓ | - | - | - | - | - |
| Image-to-Image Translation | PSNR | ✓ | - | ✓ | ✓ | ✓ | - |
| | SSIN | ✓ | - | ✓ | ✓ | ✓ | - |

*Example of implementation.* The library's usage follows intuitive patterns across all task domains. For instance, segmentation evaluation requires simple instantiation of ground truth and prediction arrays, followed by direct metric computation:

```
from allmetrics.segmentation import iou_score
y_true = [[1, 0], [1, 1]]  # Ground truth mask
y_pred = [[1, 0], [0, 1]]  # Predicted mask
iou = iou_score(y_true, y_pred)
print("IoU is: ", iou)
```

## 3. EVALUATION

We conducted a series of experiments to validate AllMetrics' performance, accuracy, and reliability in evaluating metrics across various ML tasks. These experiments included benchmarking against existing libraries, assessing computational efficiency, and testing the library's robustness under different scenarios. Below, we outline the methodology, results, and key insights from our evaluation.

### 3.1 Experimental setup

We conducted a series of experiments to evaluate the performance, accuracy, and reliability of AllMetrics by comparing it with popular frameworks such as Scikit-Learn, PyTorch, TensorFlow, PyCM, and Scikit-Image. These comparisons spanned multiple tasks and datasets covering key domains in ML: regression, classification, clustering, segmentation, and image-to-image translation. The datasets employed in our evaluation span a diverse range of ML tasks and include the following: for regression tasks, we used the Weather Dataset, House Price Dataset, PET/CT-Head and Neck Cancer Dataset; classification experiments were conducted on the Iris Dataset, Heart Disease Dataset, Breast-Cancer, Ultrasound-MRI, Online News Popularity and Car Evaluation Dataset; clustering evaluations utilized Wine-Quality Dataset, Heart Disease Dataset, Breast-Cancer, Online News Popularity Dataset, Car Evaluation Dataset, and Room Occupancy Dataset; segmentation analyses were performed using PET/CT Lung Cancer, LungCT-Diagnosis, NSCLC-Radiogenomics, QIN LUNG CT, Lung-Fused-CT-Pathology, RIDER, SPIE-AAPM and TCGA; finally, image-to-image translation was evaluated using 2D ultrasound images paired with their corresponding synthetic MRI slices for prostate cancer. For each task, we selected standard models commonly used in the literature, including decision trees for classification, linear regression for regression, k-means for clustering, 3D Attention U-Net for segmentation, and Pix2Pix for image-to-image translation. All comparative metrics were computed using each library's default parameters to reflect real-world usage scenarios where users might not perform parameter tuning. This approach also isolates the impact of algorithmic differences between implementations rather than parameter effects. The datasets used in our experiments are summarized in Table 2. Full dataset details are prepared in the supplemental File, Table 1.

**Table 2**. Summary of tasks and corresponding datasets used in this study. The table presents a comprehensive list of benchmark datasets organized by ML task type, including regression, classification, clustering, segmentation, and image-to-image translation tasks.

| Task | Dataset |
|---|---|
| Regression | Weather Dataset, House Dataset, CT/PET Head-and-Neck Cancer. |
| Classification | Iris Dataset, Heart Disease Dataset, Breast-Cancer, Ultrasound-MRI, Online News Popularity, Car Evaluation Dataset |
| Clustering | Wine-Quality Dataset, Heart Disease Dataset, Breast-Cancer, Online News Popularity Dataset, Car Evaluation Dataset, Room Occupancy Dataset |
| Segmentation | LCTSC (Lung CT Segmentation Challenge), LungCT-Diagnosis, NSCLC-Radiogenomics, QIN LUNG CT, Lung-Fused-CT-Pathology, RIDER, SPIE-AAPM Lung Challenge, TCGA |
| Image-to-Image Translation | Ultrasound-MRI |

## 3.2 Benchmark results

For benchmarking purposes, we compared AllMetrics V.1.0.0 with popular libraries such as Scikit-Learn V.1.6.1, PyTorch V.2.6.0 [10], TensorFlow V.2.18.0, PyCM V.4.3, MONAI V.1.4.0, MedPy V.0.5.2, TorchMetrics V.1.7.1, and Scikit-Image V.0.25.2. Specifically, we evaluated the Accuracy, Efficiency, and Robustness aspects. All experiments were conducted on a Windows 11 system equipped with an 11th Generation Intel Core i7-11370H processor (4 cores, 8 threads, 3.30 GHz base frequency) and 16GB of DDR4 RAM.

**3.2.1 Accuracy & Reliability**. Building on recent findings that exposed significant ID and RD in metric evaluation [1], we systematically compared AllMetrics with existing libraries across five fundamental tasks: regression, classification, clustering, segmentation, and image-to-image translation. Our analysis demonstrates that AllMetrics successfully resolves these inconsistencies through standardized implementations that reconcile ID and configurable reporting options that eliminate RD.

*In Classification Tasks*. The discrepancies observed across different ML libraries in classification tasks primarily stem from RD — variations in how metric results are documented and presented. As shown in Tables 3 and 4, these inconsistencies are particularly evident in binary and multi-class classification scenarios. Table 3 presents results from the Heart Disease dataset, while Table 4 illustrates findings from the Iris dataset, highlighting differences in metric reporting across frameworks. In binary classification (see Table 3), AllMetrics follows the same behavior as PyCM, reporting metric values for each class separately (e.g., {0: 0.86, 1: 0.76} for Precision). In contrast, most other frameworks report a single aggregated value, typically using micro-averaging or macro-averaging strategies. This difference in output format can lead to misinterpretation when comparing results across platforms or attempting to reproduce findings. Similarly, in multi-class classification (see Table 4), AllMetrics maintains consistency by returning per-class results by default, enabling users to inspect performance at a granular level. Other libraries often return different results, for example, TensorFlow returns micro (e.g., 0.95 for Precision), and most of them give a single average score (e.g., 0.65 for Precision), which may obscure important variations between individual classes. Full classification results for other datasets/tasks/metrics are provided in Supplemental File, Tables 2-21.

**Table 3**. Comparative evaluation of classification metrics across different libraries for heart disease prediction .This table compares precision, recall, F1 Score, and Jaccard Index values obtained from multiple evaluation libraries (AllMetrics, Caret, Yardstick, PyCM, and other tools) for binary classification of heart disease. Values are shown separately for class 0 (non-disease) and class 1 (disease), demonstrating variations in metric calculations across different analytical frameworks. As evident from the results, AllMetrics and PyCM yield identical outputs for all metrics, suggesting consistent algorithmic implementations between these two libraries.

| Metric | Library | Value |
|---|---|---|
| Precision | AllMetrics | {0: 0.86, 1: 0.76} |
| | Caret & Yardstik | 0.86 |
| | PyCM | {0: 0.86, 1: 0.76} |
| | Other | 0.76 |
| Recall | AllMetrics | {0: 0.75, 1: 0.87} |
| | Caret & Yardstik | 0.75 |
| | PyCM | {0: 0.75, 1: 0.87} |
| | Other | 0.87 |
| F1 Score | AllMetrics | {0: 0.80, 1: 0.82} |
| | Caret & Yardstik | 0.80 |
| | PyCM | {0: 0.80, 1: 0.82} |
| | Other | 0.82 |
| Jaccard Index | AllMetrics | {0: 0.66, 1: 0.68} |
| | TensorFlow | 0.67 |
| | PyCM | {0: 0.66, 1: 0.68} |
| | Other | 0.68 |

**Table 4**. Comparative evaluation of multi-class classification metrics across different libraries for heart disease prediction. This table represents precision, recall, F1 Score, Matthews Correlation Coefficient (MCC), and Jaccard Index values obtained from various evaluation libraries (AllMetrics, TensorFlow, PyCM, and other tools) for multi-class classification on the Iris dataset. Values are shown separately for each class (0, 1, and 2), highlighting performance variations across different species classifications. The results demonstrate that AllMetrics and PyCM produce identical outputs for all metrics, indicating consistent computational approaches between these libraries.

| Metric | Library | Value |
|---|---|---|
| Precision | AllMetrics | {0: 1.0, 1: 0.4, 2: 0.55} |
| | TensorFlow | 0.95 |
| | PyCM | {0: 1.0, 1: 0.4, 2: 0.55} |
| | Other | 0.65 |
| Recall | AllMetrics | {0: 0.90, 1: 0.44, 2: 0.55} |
| | TensorFlow | 1.00 |
| | PyCM | {0: 0.90, 1: 0.44, 2: 0.55} |
| | Other | 0.63 |
| F1 Score | AllMetrics | {0: 0.95, 1: 0.421, 2: 0.55} |
| | TensorFlow | 0.98 |
| | PyCM | {0: 0.95, 1: 0.421, 2: 0.55} |
| | Other | 0.64 |
| MCC | AllMetrics | {0: 0.93, 1: 0.15, 2: 0.28} |
| | PyCM | {0: 0.93, 1: 0.15, 2: 0.28} |
| | Other | 0.45 |
| Jaccard Index | AllMetrics | {0: 0.9, 1: 0.27, 2: 0.37} |
| | PyCM | {0: 0.9, 1: 0.27, 2: 0.37} |
| | Other | 0.51 |

*In Regression Tasks*. Our evaluation of regression metrics, particularly the $R^2$ coefficient, demonstrates strong alignment between AllMetrics and well-established libraries such as Scikit-Learn, Matlab, and TorchMetrics. As shown in Table 5, which presents results from the Weather Dataset, the values obtained using AllMetrics are exactly identical to those produced by Scikit-Learn and Matlab across all tested cases. Furthermore, when compared with TorchMetrics, the outputs match up to seven decimal places, confirming its numerical precision and consistency.

**Table 5**. Comparative evaluation of regression R Squared ($R^2$) metric across libraries AllMetrics, TorchMetrics, Scikit-Learn, and Matlab for heart disease prediction. The results demonstrate that AllMetrics, Scikit-Learn, and Matlab produce identical outputs for this metric, indicating consistent computational approaches between these libraries.

| Metric | Library | Value |
|---|---|---|
| R Squared ($R^2$) | AllMetrics | 0.772177853490543 |
| | TorchMetrics | 0.772177875041962 |
| | Scikit-Learn | 0.772177853490543 |
| | Matlab | 0.772177853490543 |

*In Clustering Tasks*. The results produced by AllMetrics demonstrate strong alignment with those of Scikit-Learn, one of the most widely used libraries in ML. Our analysis shows that the core clustering metrics—such as Rand Score, Silhouette Score, and Davies-Bouldin Index—are computed in a manner consistent with Scikit-Learn's implementations. This similarity not only validates the correctness of AllMetrics' metric calculations but also ensures compatibility with existing workflows.

*In Segmentation Tasks*. For segmentation evaluation, we utilize a validated collection of 240 lung CT tumor segmentations sourced from 8 independent multi-center datasets. As shown in Table 6, the results demonstrate that AllMetrics produces segmentation scores closely aligned with those obtained using MedPy, a well-established library for medical image analysis. This similarity confirms the accuracy and reliability of our implementation while ensuring compatibility with existing tools used in clinical and research settings.

**Table 6**. Comparative evaluation of Segmentation Hausdorff Distance metric across libraries AllMetrics, MONAI, and MedPy for Lung CT tumor. The results demonstrate that AllMetrics and MedPy produce identical outputs for this metric, indicating consistent computational approaches between these libraries.

| Metric | Library | Value |
|---|---|---|
| Hausdorff Distance | AllMetrics | 35.94 |
| | MONAI | 22.43 |
| | MedPy | 35.94 |

*In Image-to-Image Translation Tasks*. We utilized a dataset comprising 2D ultrasound images and their corresponding synthetic MRI slices for prostate cancer, generated using a Pix2Pix-based image-to-image translation model. The dataset includes data from 24 patients, with a total of 1409 image pairs of size 128×128 pixels. In this setup, the synthetic MRIs generated by the model are treated as predictions, while the real MRIs serve as ground truth for evaluation purposes. As shown in Table 7, the results across different libraries exhibit notable variation, particularly in metrics such as SSIM, which are sensitive to implementation details. For instance, some libraries apply fixed window sizes during SSIM computation, limiting flexibility and potentially affecting result interpretation.

**Table 6**. Comparative evaluation of image-to-image translation structural similarity index (SSIM) metric across libraries AllMetrics, Scikit-Image, Pytorch, and TensorFlow for 2D ultrasound images. The results demonstrate that AllMetrics and MedPy produce identical outputs for this metric, indicating consistent computational approaches between these libraries. The differences in window size selection have led to varying results. Here, PyTorch and TensorFlow produce identical outputs because they use the same window size configuration, while AllMetrics employs a user-adjustable parameter for custom window size selection.

| Metric | Library | Value |
|---|---|---|
| SSIM | AllMetrics | 0.14 |
| | Scikit-Image | 0.11 |
| | PyTorch | 0.13 |
| | TensorFlow | 0.13 |

*Efficiency.* AllMetrics demonstrates superior computational efficiency compared to existing libraries. In image-to-image translation tasks evaluating SSIM on the Prostate Cancer dataset, AllMetrics achieved the second fastest computation time among four benchmarked libraries (Figure 1). Complete results across all tasks and metrics are presented in Supplemental Files, Tables 22-41.

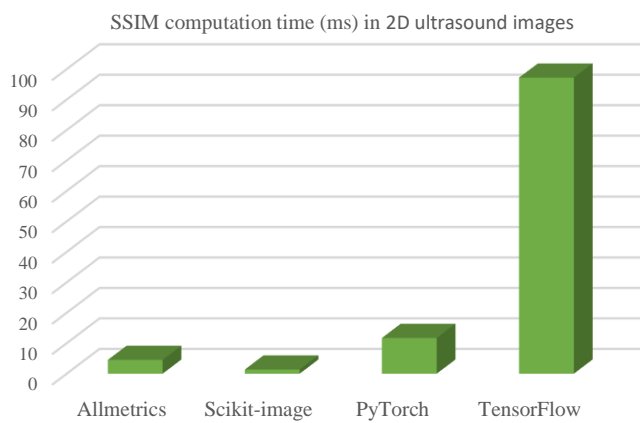

**Figure 1**: Computation time (ms) comparison for structural similarity index (SSIM) evaluation on the Prostate Cancer dataset across four libraries (lower is better). AllMetrics demonstrates second-best performance in this benchmark.

*Robustness.* Testing the library's ability to handle edge cases - including empty masks, class imbalance, and mismatched dimensions - demonstrates its strength in ensuring data integrity. AllMetrics excels in addressing these edge cases through comprehensive validation mechanisms. For segmentation tasks, it successfully detects and warns about empty masks, while for classification tasks, it flags severe class imbalances with actionable warnings. These validation checks ensure reliable metric computation across diverse scenarios while maintaining computational efficiency.

### 3.3 Limitations and future work

While AllMetrics demonstrates strong performance across a wide range of tasks, there are areas for improvement:

*Scalability.* For extremely large datasets (e.g., high-resolution satellite imagery), further optimizations may be needed to reduce memory usage.

*Metric Coverage.* Although AllMetrics V.1.0.0 supports a broad set of metrics, additional metrics for current tasks and other specialized tasks (e.g., anomaly detection) could be incorporated in future versions.

Through these efforts, AllMetrics aims to solidify its role as a foundational tool for reliable model evaluation in the ML community.

## 4. DISCUSSION

The ML community faces significant reproducibility challenges stemming from inconsistent implementations and reporting of evaluation metrics across different platforms. These inconsistencies manifest as two distinct but interrelated problems: ID, where the same metric is computed using divergent algorithms across libraries, and RD, where results are presented using incompatible formats or aggregations. Research has demonstrated that these discrepancies substantially hinder reliable model comparison and experimental replication [1].

A concrete examination of common classification metrics reveals the scope of this problem. Metrics including Precision, Recall, F1 Score, Balanced Accuracy, Jaccard Index, and Matthews Correlation Coefficient (MCC) exhibit notable variations across major ML frameworks. TensorFlow and TorchMetrics implement micro-averaging for Precision and Recall computations, while alternative libraries typically provide class-wise results. These inconsistencies create substantial barriers to meaningful performance comparison and reliable knowledge dissemination.

AllMetrics V.1.0.0 addresses these challenges through systematically standardized metric implementations with configurable computation parameters. The library's design specifically targets the resolution of both ID and RD issues through several key mechanisms:

For classification tasks, AllMetrics introduces an explicit average parameter that provides controlled aggregation options:

*Macro*: Computes metrics independently per class, followed by the unweighted mean
*Micro*: Aggregates contributions across all classes before final computation
*Weighted*: Applies class-frequency weighted means to per-class metrics
*None* (Class-wise output): Returns unevaluated metrics for each class (default)

This flexible approach is complemented by comprehensive documentation accessible through the "*get_metric_details()*" function, which provides complete parameter specifications and usage guidelines for every metric. This parameterized design effectively resolves reporting discrepancies while maintaining implementation transparency.

Regression analysis benefits from similar standardization, particularly for the R² metric, where implementation variance is well-documented. AllMetrics supports both traditional linear R² (matching Scikit-Learn and TorchMetrics) and adjusted R² formulations (aligned with StatsModels [11] and Yardstick [12]), with explicit parameter control over the computation method. This dual support ensures compatibility with diverse analytical requirements while eliminating ambiguity in result interpretation.

In segmentation tasks, the library's approach to computing spatial similarity metrics—such as Dice Score, IoU, and Hausdorff Distance—stands out for its mathematical rigor and robustness. Unlike MONAI's implementation, which relies on distance transform methods to estimate surface-wise distances between predicted and ground-truth segmentations, our method (in line with MedPy) directly computes Euclidean distances between foreground boundary points. While MONAI's approach can be sensitive to image resolution and surface smoothness, our method provides a more geometrically accurate and interpretable distance measure, especially in sparse or noisy segmentation tasks. Moreover, our implementation explicitly supports partial C distance (e.g., HD95), which is commonly used in clinical evaluation benchmarks. It also handles degenerate cases, such as empty masks, with consistent behavior—returning ∞ rather than undefined or zero values. This level of detail ensures reliable and meaningful comparisons in real-world medical imaging applications where edge cases are common.

In image-to-image translation tasks, the computation of SSIM reveals notable differences in implementation across popular frameworks. Specifically, TensorFlow and TorchMetrics utilize a fixed window size of 11×11 pixels for SSIM calculation, which is not adjustable by the user. This rigid configuration limits flexibility and may not be optimal for all applications, especially those involving high-resolution or domain-specific imagery where spatial context varies significantly.

By contrast, AllMetrics adopts the same default window size as Scikit-Image, namely 7×7 pixels, while also introducing a configurable parameter window_size. This feature empowers users to dynamically adjust the window dimensions according to their specific application requirements. For instance, in medical imaging or satellite data analysis, where fine-grained structural details are critical, a smaller window may better capture localized variations, whereas larger windows might be preferred for global structure preservation in panoramic views.

Beyond metric standardization, AllMetrics incorporates rigorous data validation protocols that proactively identify and address input quality issues. These mechanisms work in concert with the standardized metric computations to deliver reliable, reproducible evaluations. The library's design philosophy emphasizes both theoretical consistency, through mathematically precise, well-documented implementations, and practical usability, via intuitive interfaces and comprehensive validation.

This dual focus on standardization and reliability positions AllMetrics as a valuable tool for addressing the ML community's growing need for reproducible evaluation frameworks. By systematically resolving both ID and RD challenges while maintaining flexibility for diverse use cases, the library advances the state of metric evaluation practice and supports more reliable research outcomes.

## 5. CONCLUSION

AllMetrics presents a comprehensive solution to the longstanding challenges of standardization and reliability in ML evaluation. By unifying task-specific metrics with rigorous data validation, the library establishes a systematic framework for assessing model performance across regression, classification, clustering, segmentation, and image-to-image translation tasks. Its modular architecture not only accommodates diverse use cases but also enforces data integrity through automated checks for common pitfalls such as empty masks, class imbalances, and dimensional inconsistencies. The library's optimized implementations, intuitive API, and open-source availability make it a practical tool for both research and production environments. By resolving implementation and reporting discrepancies inherent in existing solutions, AllMetrics enhances the reproducibility and comparability of ML evaluations—a critical advancement as the field increasingly prioritizes rigorous and transparent benchmarking.

**DATA AND CODE AVAILABILITY/SUPPLEMENTAL FILE**. The AllMetrics library is open-source and freely available on GitHub. The supplemental file and source code, along with detailed documentation and usage examples, can be accessed at the following repository:
*https://github.com/MohammadRSalmanpour/allmetrics-A-Unified-Python-Library-for-Standardized-Metric-Evaluation-in-Machine-Learning*

**ACKNOWLEDGMENT.** This study was supported by Virtual Collaboration (VirCollab, *www.vircollab.com*) Group as well as Technological Virtual Collaboration Corporation (TECVICO CORP.), Vancouver, Canada. This work was also supported by the Canadian Foundation for Innovation-John R. Evans Leaders Fund (CFI-JELF) program [Grand ID 42816]. We acknowledge the support of the Natural Sciences and Engineering Research Council of Canada (NSERC), [RGPIN-2023-03575] and Discovery Horizons Grant [DH-2025-00119]. Cette recherche a été financée par le Conseil de recherches en sciences naturelles et en génie du Canada (CRSNG), [RGPIN-2023-0357].

**CONFLICT OF INTEREST**. The co-authors Mehrdad Oveisi, Sonya Falahati, and Mohammad R. Salmanpour are affiliated with TECVICO CORP. The remaining co-authors declare no relevant conflicts of interest.